\documentclass[british,format=sigconf]{aamas}
\usepackage[latin9]{inputenc}
\usepackage{color}
\usepackage{array}
\usepackage{booktabs}
\usepackage{multirow}
\usepackage{graphicx}

\makeatletter

\providecommand{\tabularnewline}{\\}

\usepackage[linesnumbered,ruled]{algorithm2e}
\usepackage{url}
\usepackage{capt-of}
\usepackage{hhline}

\setcopyright{ifaamas}
\acmConference[AAMAS '22]{Proc.\@ of the 21st International Conference
on Autonomous Agents and Multiagent Systems (AAMAS 2022)}{May 9--13, 2022}
{Online}{P.~Faliszewski, V.~Mascardi, C.~Pelachaud,
M.E.~Taylor (eds.)}
\copyrightyear{2022}
\acmYear{2022}
\acmDOI{}
\acmPrice{}
\acmISBN{}

\author{Dung Nguyen}
\affiliation{
  \institution{A$^2$I$^2$, Deakin University}
  \city{Geelong}
  \country{Australia}}
\email{dung.nguyen@deakin.edu.au}

\author{Phuoc Nguyen}
\affiliation{
  \institution{A$^2$I$^2$, Deakin University}
  \city{Geelong}
  \country{Australia}}
\email{phuoc.nguyen@deakin.edu.au}

\author{Hung Le}
\affiliation{
  \institution{A$^2$I$^2$, Deakin University}
  \city{Geelong}
  \country{Australia}}
\email{thai.le@deakin.edu.au}

\author{Kien Do}
\affiliation{
  \institution{A$^2$I$^2$, Deakin University}
  \city{Geelong}
  \country{Australia}}
\email{k.do@deakin.edu.au}

\author{Svetha Venkatesh}
\affiliation{
  \institution{A$^2$I$^2$, Deakin University}
  \city{Geelong}
  \country{Australia}}
\email{svetha.venkatesh@deakin.edu.au}

\author{Truyen Tran}
\affiliation{
  \institution{A$^2$I$^2$, Deakin University}
  \city{Geelong}
  \country{Australia}}
\email{truyen.tran@deakin.edu.au}

\acmSubmissionID{438}

\newcommand{\BibTeX}{\rm B\kern-.05em{\sc i\kern-.025em b}\kern-.08em\TeX}


\renewcommand{\citet}{\citep}
\usepackage{balance}

\makeatother

\usepackage{babel}
\begin{document}
\pagestyle{fancy} 
\fancyhead{}
\title{Learning Theory of Mind via Dynamic Traits Attribution}
\begin{abstract}
Machine learning of Theory of Mind (ToM) is essential to build social
agents that co-live with humans and other agents. This capacity, once
acquired, will help machines infer the mental states of others from
observed contextual action trajectories, enabling future prediction
of goals, intention, actions and successor representations. The underlying
mechanism for such a prediction remains unclear, however. Inspired
by the observation that humans often infer the character traits of
others, then use it to explain behaviour, we propose a new neural
ToM architecture that learns to generate a latent trait vector of
an actor from the past trajectories. This trait vector then multiplicatively
modulates the prediction mechanism via a `fast weights' scheme in
the prediction neural network, which reads the current context and
predicts the behaviour. We empirically show that the fast weights
provide a good inductive bias to model the character traits of agents
and hence improves mindreading ability. On the indirect assessment
of false-belief understanding, the new ToM model enables more efficient
helping behaviours. 

\end{abstract}
\maketitle

\section{Introduction}

The capacity of a social agent to predict and interpret the behaviours
of others is essential for it to thrive. A basis for this mindreading,
also known as theory of mind (ToM), is the ability to attribute \emph{transient}
mental states -- knowledge, emotions, beliefs, desires and intention
-- to others and use the attribution to reason about their actions
\citet{gallese1998mirror,gordon1986folk,gopnik1992child,premack1978does,rusch2020theory}.
We also often attribute \emph{stable} character-traits to others in
order to interpret their behaviours \citet{ames2011impression,kelley1967attribution,meltzoff2013learning}.
While these two distinct attribution skills have been known to be
related, it remains unclear how they are integrated into a coherent
system. A recent theory has been pushed forward, suggesting that temporally
stable character traits can be used to generate a prior probability
distribution for hypotheses about mental states \citet{westra2018character}.
A computational theory to realise the hypothesis remains open.

Inspired by theories of human theory of mind, we seek to embed a learning
theory of mind system inside a social agent to predict the behaviours
of others. The system makes minimal assumptions about the underlying
mental structure of the other agents, but instead learns to construct
the latent character traits by observing their past behaviours and
infers the mental states from the current behaviour. We wish to design
the system so that the traits drive the predictive mechanism from
states to actions/behaviours. This is unlike traditional mindreading
in AI which studies symbolic plan and goal recognition \citet{geib2009probabilistic,kautz1986generalized,mirsky2021introduction,sohrabi2016plan}.
This approach makes strong assumptions about, and requires detail
descriptions of, the domain. The Bayesian approach to ToM (BToM) \citet{baker2011bayesian,baker2017rational,wen2019probabilistic}
relies on the bounded rationality assumption, i.e., actors will maximise
their own utility based on partial observations, to build a model
of others. Here the past information about the goal of actor serves
as a prior distribution to update the model. This treatment could
not cover the case when complex past behaviours in different environments
maintain information about the stable mental state (traits) of actors.
BToM focuses on analysing the current behaviours of the other, but
does not explain how to incorporate the individuality expressed through
past behaviours into executing prediction. 

The deep learning approach to ToM has recently been brought forward
to leverage the computational efficiency and architectural flexibility
of neural networks \citet{oguntola2021deep,rabinowitz2018machine,moreno2021neural}.
A model proposed in \citet{rabinowitz2018machine} called Theory of
Mind neural network (ToMnet) jointly models prior actor characters
and current mental states from observations. In particular, ToMnet
constructs a character embedding from past behaviours using a character
network, and a mental embedding of the current trajectory of an actor
using a mental network. The two embedding vectors are then combined
as input to a prediction network to infer the actor's goal and future
behaviours. This architecture is trained with a large amount of data
sampled from a mixed population of actors. However, it remains unclear
whether ToMnet could learn to provide good predictions after being
trained in more realistic settings, such as when it could only observe
one type of actor in a time period. The work in \citet{oguntola2021deep}
focuses on the interpretability of the ToM models. 

\begin{figure*}
\begin{centering}
\includegraphics[width=0.6\textwidth]{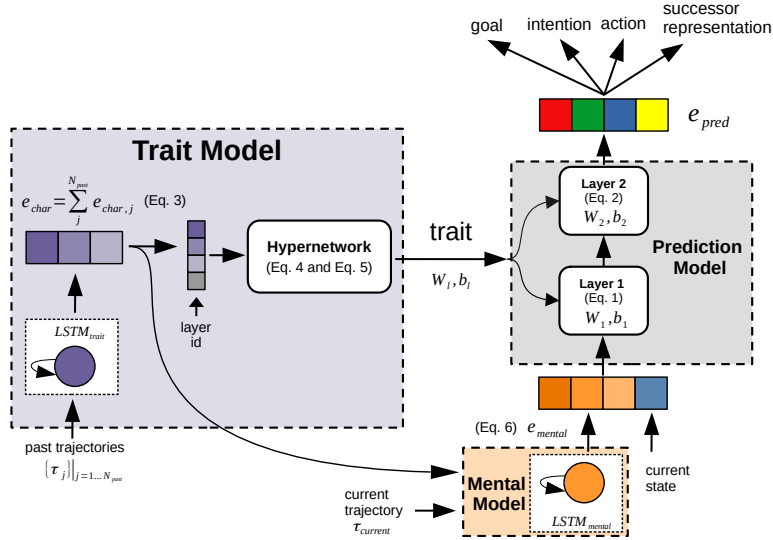}
\par\end{centering}
\caption{\label{fig_hyper_tom_with_intention} Trait-based Theory of Mind (Trait-ToM)
architecture. Our architecture generates the trait of an actor based
on its historical behaviours. The trait then modulates the prediction
path from the mental and environmental states to the future behaviours.
}
\end{figure*}

Different from these works, we employ the \emph{fast weight} concept
\citet{ba2016using,schmidhuber1992learning} to represent the character
traits of actors in a behaviour prediction network. Unlike standard
slow weights which are fixed after training, fast weights are computed
on-the-fly at inference time, conditioned on the observations. More
specifically, these fast weights are generated by a hypernetwork \citet{ha2016hypernetworks,jayakumar2020multiplicative}
using past behaviours. We use these character trait weights to modulate
the mind prediction in a multiplicative manner. A ToM observer will
have the freedom to change the weights of the prediction networks
independently for each character trait. We show in the experiments
that this capacity helps the ToM observer correctly predict the actor's
goal, intention, and trajectories and perform better in complex tasks.
To the best of our knowledge, our paper is the first work implementing
and analysing the idea of fast weights and hypernetworks for mindreading
tasks. Our main contribution is introducing a new type of Trait-based
ToM (Trait-ToM) agent (the observer), which can represent the character
traits of other actors and use it to make behaviour predictions of
others. We verify the predictive power of Trait-ToM on a suite of
tasks in a key-door environment with a mixed population of actors
and different realistic training settings and demonstrate promising
results.

\section{Problem Formulation}

We consider a family of partially observable Markov decision processes
(POMDPs) $\mathcal{W}=\cup_{j}\mathcal{W}_{j}$, where each environment
is a tuple $\mathcal{W}_{j}=\left\langle S_{j},A_{j},T_{j}\right\rangle $
of the state space $S_{j}$, the action space $A_{j}$ and the transition
functions $T_{j}$. Acting on the environments is a family of actors
$\mathcal{A}=\cup_{i}\mathcal{A}_{i}$, each of which has its own
observation space $\mathcal{O}_{i}$, observation function $\Omega_{i}:\mathcal{O}_{i}\times\mathcal{S}_{j}\times A_{j}\longmapsto\left[0,1\right]$,
reward function $R_{i}$, and a policy $\pi_{i}$, i.e. $\mathcal{A}_{i}=\left\langle \mathcal{O}_{i},\Omega_{i},R_{i},\mathcal{W}_{i},\pi_{i}\right\rangle $.
In the simplest form, the\emph{ type} of each actor is defined by
its perception ability, preferences, and strategy.  Finally, we
consider an observer who can observe the \emph{behaviours} of the
actors. Here, the behaviour of the actor $i$ in the environment $W_{j}$
is represented by the trajectory $\tau_{ij}=\left(s^{(t)},a^{(t)}\right)_{t=0}^{T-1}$
with $s^{(t)}\in S_{j}$, $a^{(t)}\in A_{j}$, and $T$ is the length
of the trajectory. 

The observer or theory of mind (ToM) agent first observes a set of
$N_{past}$ \emph{past trajectories} $\left\{ \tau_{ij}\right\} _{1}^{N_{past}}$
of an actor $i$ in different environments $\mathcal{W}_{j}$ with
$j=1,\dots,N_{past}$. We hypothesise that these past behaviours exhibit
the character of this actor, allowing the formation of a good prior
for predicting its behaviours. We then ask the ToM agent to predict
the behaviours of this actor in the \emph{current environment} by
observing the \emph{current trajectory}, including the state and action
pairs up to the query time step, $\tau_{i,current}=\left(s_{i,current}^{(t)},a_{i,current}^{(t)}\right)_{t=0}^{T_{q}-1}$
and the \emph{current state} of the world $s_{i,current}^{(T_{q})}$.
Here, $T_{q}$ is the time step that the ToM agent is queried to make
prediction. The behaviour of actors that we would like our model to
answer includes preferences, one step ahead actions, intentions and
future visit state (via successor representations). To reduce the
notation load, we will drop the $i$ notion if there is no confusion.
In this paper, we use \emph{the observer} and \emph{the ToM agent}
interchangeably; \emph{the actor} refers to the observed agent.

\section{Method}

\subsection{Trait-based Theory of Mind Architecture}

In this section, we introduce the architecture of Trait-based Theory
of Mind (Trait-ToM) model for the observer. There are three modules:
(1) Prediction Model, (2) Trait Model, and (3) Mental Model. In
this architecture, the Trait Model captures the long-term trait of
actors in the past trajectories while the Mental Model represents
the recent behaviours in the current trajectory. The outputs of the
two modules will be used by the prediction module through our proposed
fast weight mechanism. In the following, we detail the operation of
each module. The architecture is shown in Figure \ref{fig_hyper_tom_with_intention}.

\paragraph{Prediction Model}

Let $s$ be the current environmental state and $\mathbf{e}_{mental}$
be the current estimated mental state vector of an actor. Based on
this information and the past behaviours, we wish to predict the four
outputs of the actor: preference, intention, action, and successor
representation. The prediction network will generate an output vector
$\mathbf{e}_{pred}$ as follows:
\begin{align}
h & =\sigma\left(\mathbf{W}_{1}f_{pred}\left(s,\mathbf{e}_{mental}\right)+\mathbf{b}_{1}\right),\,\text{and}\label{eq:PredNet1}\\
\mathbf{e}_{pred} & =\mathbf{W}_{2}h+\mathbf{b}_{2}.\label{eq:PredNet2}
\end{align}
for some feature extractor $f_{pred}\left(\cdot\right)$ which is
a neural network, activation function $\sigma\left(\cdot\right)$,
and weights $\mathbf{W}_{1}$, $\mathbf{b}_{1}$, $\mathbf{W}_{2}$,
and $\mathbf{b}_{2}$. The mental state $\mathbf{e}_{mental}$ is
generated from the mental model in Eq.~(\ref{eq:Mental-State}).
The output vector $\mathbf{e}_{pred}$ then serves as input for four
prediction heads, which are are goal, intention, action, and successor
representation.

In Eqs.~(\ref{eq:PredNet1},\ref{eq:PredNet2}), the key to our formulation
is that the fast weights $\left\{ \mathbf{W}_{l},\mathbf{b}_{l}:l\in\{1,2\}\right\} $
represent the \emph{individual characteristics }or\emph{ the character
traits}, which are \emph{functions} of the past behaviours. Unlike
ToMnet, which uses a vector to represent the trait as the input to
the prediction net, our fast-weight traits are higher in capacity
and directly modulate the function of the prediction net through multiplicative
mechanisms. These weights are computed by our trait model, which we
present next in Eqs.~(\ref{eq:Hyper-W},\ref{eq:Hyper-b}). In other
words, \emph{the traits modulate the prediction path} from the current
mental and environmental states to the outcomes.

\paragraph{Trait Model}

The trait network takes past trajectories of an actor in $N_{past}$
environment $\left\{ \tau_{j}\right\} _{j=1}^{N_{past}}$ as inputs
and generates the trait vector of the actor. For each past trajectory
$j$, it maintains a dynamic state vector at each time step $t$ by
a long short-term memory network \citet{hochreiter1997long} as follows:

\[
h_{j}^{(t)}=\mathrm{LSTM}_{trait}\left(\mathbf{x}_{j}^{(t)},h_{j}^{(t-1)}\right),
\]
where $\mathbf{x}_{j}^{(t)}$ is the features extracted from the state-action
pair using a neural network, i.e., $\mathbf{x}_{j}^{(t)}=f_{trait}(s_{j}^{(t)},a^{(t)})$.
The trait embedding vector is computed by averaging over all the past
trajectories:

\begin{eqnarray}
\mathbf{e}_{char} & =\frac{1}{N_{past}} & \sum_{j=1}^{N_{past}}\mathrm{ReLU}\left(\mathrm{Linear}\left(h_{j}^{(T_{j})}\right)\right).\label{eq:Char-Embedding}
\end{eqnarray}

To directly modulate the prediction path and create the multiplicative
interaction between past behaviours and the mental embedding, we use
a hypernetwork which takes $\mathbf{e}_{char}$ as input to generate
the weights $\mathbf{W}_{l}$ and biases $\mathbf{b}_{l}$, $l\in\{1,2\}$,
of the prediction network in Eqs.~(\ref{eq:PredNet1},\ref{eq:PredNet2}): 

\begin{align}
\mathbf{W}_{l} & =\sigma\left(\mathrm{Linear}\left(\left[\mathbf{e}_{char},\mathbf{c}_{l}\right]\right)\right),\,\text{and}\label{eq:Hyper-W}\\
\mathbf{b}_{l} & =\sigma\left(\mathrm{Linear}\left(\left[\mathbf{e}_{char},\mathbf{c}_{l}\right]\right)\right),\,l\in\{1,2\}\label{eq:Hyper-b}
\end{align}
where the one-hot vector $\mathbf{c}_{l}=\text{onehot}(l)$ is used
as an additional input to indicate which layer $l$ to generate weights.
The set of weights $\left\{ \mathbf{W}_{l},\mathbf{b}_{l}:l\in\{1,2\}\right\} $
serves as the\emph{ representation of dynamic traits of the actor,}
 which changes whenever the behaviour history is updated. We hypothesise
that the dynamic traits is critical to capture diverse behaviours
of multi-agents. Each agent should triggers a signature fast weight
to determine the prediction for its future behaviour.

\paragraph{Mental Model}

The mental model reads the current trajectory and estimates the mental
state using the following dynamics: 

\[
h_{mental}^{(t)}=\mathrm{LSTM}_{mental}\left(\left[\mathbf{x}_{mental}^{(t)},\mathbf{e}_{char}\right],h_{mental}^{(t-1)}\right),
\]
where $\mathbf{e}_{char}$ is the trait embedding computed in Eq.~(\ref{eq:Char-Embedding}),
and $\mathbf{x}_{mental}^{(t)}$ denotes features extracted from the
current state-action pair, i.e., $\mathbf{x}_{mental}^{(t)}=f_{mental}(s_{current}^{(t)},a^{(t)})$.
The initial hidden state is computed from the trait embedding $h_{mental}^{(0)}=\mathrm{Linear}\left(\mathbf{e}_{char}\right)$.
The mental embedding vector is computed as: 

\begin{equation}
\mathbf{e}_{mental}=\mathrm{ReLU}\left(\mathrm{Linear}\left(h_{mental}^{(t)}\right)\right).\label{eq:Mental-State}
\end{equation}
It serves as an input for the prediction network in Eq.~(\ref{eq:PredNet1}).

\subsection{Loss functions}

We feed $e_{pred}$ computed from Eq.~(\ref{eq:PredNet2}) to different
heads corresponding to different targets that we want to predict.
To train the Trait-based ToM, we use the following losses:

\[
\mathcal{L}=\mathcal{L}_{pref}+\mathcal{L}_{intention}+\mathcal{L}_{action}+\mathcal{L}_{SR}.
\]
The four component losses are as follows:

\paragraph{Preference Prediction}

Each actor has its own preference, e.g., a colour. The negative log-likelihood
of the preference of the actor is therefore:

\[
\mathcal{L}_{pref}=\sum_{pref}-\log p\left(pref\left|\mathbf{e}_{pred}\right.\right),
\]
where the $p_{pref}$ is modelled by a neural network that takes $\mathbf{e}_{pred}$
as its input.

\paragraph{Intention Prediction}

In our setting, at each time step, the actor maintains a sub-plan
(intention) such as going to a place or finding objects. We record
the intention of actors at every time step, and compute the negative
log-likelihood of the intention of the actor: 

\[
\mathcal{L}_{intention}=\sum-\log p\left(intent\left|\mathbf{e}_{pred}\right.\right).
\]

\paragraph{Action Prediction}

We use the negative log-likelihood of the true action of the actor:

\[
\mathcal{L}_{action}=-\log\pi\left(a_{t}\left|\mathbf{e}_{pred}\right.\right).
\]

\paragraph{Successor Representation Prediction}

We use an empirical successor representation \citet{dayan1993improving}
(SR) to compute the SR loss

\[
\mathcal{L}_{SR}=\sum_{\gamma_{SR}}\sum_{s}-SR_{\gamma_{SR}}\left(s\right)\log\tilde{SR}_{\gamma_{SR}}\left(s\right),\,\text{with}
\]

\[
SR_{\gamma}\left(s\right)=\frac{1}{Z}\sum_{t'=0}^{T-t}\gamma_{SR}^{t'}I\left(s_{t+ t'}=s\right)
\]
where $T$ is the episode length, $t$ is the time at which the successor
representation is computed, $Z$ is a normalisation constant, $\gamma_{SR}\in\left(0,1\right)$
is the discount factor and $I\left(s_{t+ t'}=s\right)$ is an
indicator function, which returns $1$ if $s_{t+ t'}=s$ and
$0$ otherwise.

\section{Case Study: Key-Door Environment}

\subsection{Experiment Settings}

\begin{figure}
\begin{centering}
\includegraphics[width=1\columnwidth]{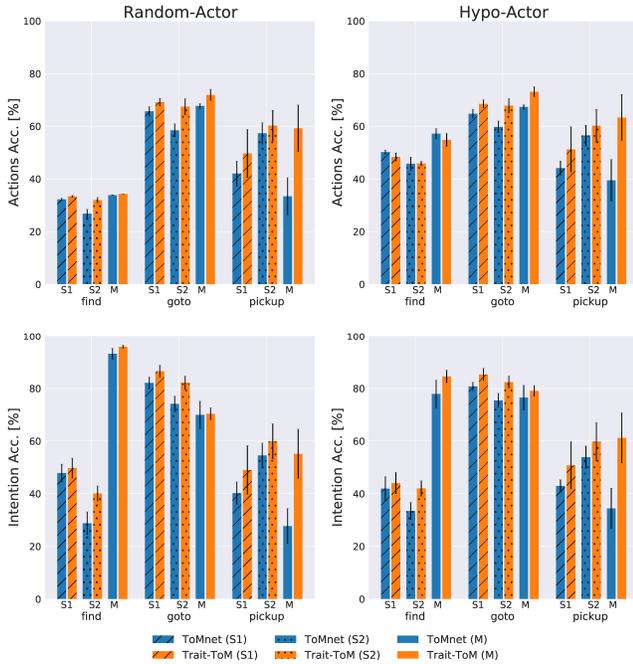} 
\par\end{centering}
\caption{\label{fig_sequence_actors_prediction}Performance of ToMnet and Trait-ToM
after being trained on mixed (M) and sequential settings (S1 and S2).
The y-axis shows the accuracies of action (top row) and intention
prediction (bottom row) on the random-actor population (left column)
and the hypo-actor population (right column), conditioned on the intention
of the actors (find, goto, pickup).}
\end{figure}

\paragraph{Environment}

In this section, we conduct experiments on the Key-Door Environment
using the \emph{gym-minigrid }framework \citep{gym_minigrid}. In
this environment, there are two types of object $\{\mathrm{key},\mathrm{door}\}$
in four different colours $\{\mathrm{red},\mathrm{green},\mathrm{blue},\mathrm{yellow}\}$.
An actor has its own preference for the colour. An episode is terminated
when the actor picks up the key and goes to the door in its preferred
colour.

\paragraph{Goal-directed Actors}

We construct the actors that have a consistent goal during one episode,
e.g. picking up the key and going to the door in the preferred colour.
We assume each actor has beliefs about the positions of all objects
in the scene, as well as the ability to memorise all visited cells.
The actor is able to detect its \emph{false belief} and update its
belief according to recent observations. At each time step, it has
an intention to either $\texttt{find()}$, $\texttt{goto()}$, or
$\texttt{pickup()}$. The actors have the belief-desires-intentions
(BDI) architecture \citep{bratman1987intention,georgeff1998belief}
and dynamically switching between three plans $\texttt{find()}$,
$\texttt{goto()}$, or $\texttt{pickup()}$ can be considered as changing
intentions. 

Actors are different in the strategy they use to execute the intention
$\texttt{find(object)}$: (1) the \emph{random-actor} that can only
take random walks to find its preferred object; and (2) the wiser
actor that maintains a hypothesis about the position of its preferred
key and door. The latter type of actor tests its hypothesis by going
to these positions and seeking for the object. If the actor cannot
find the object there, it will make a new guess about the object position.
For this reason, we call this actor a \emph{hypo-actor} (shorthand
for \emph{hypothesis testing actor}). In this paper, we only explore
the salient traits (e.g. \emph{smart - not smart} or \emph{hypo -
random})~\citep{rosati2001theory} and leave other traits, e.g. finding
constraints on the value of information and risk aversion, for future
work. Each actor partially observes a square area which indicates
the \emph{field of view} (FoV). In sum, there are $32$ actors which
are characterised by combinations of $4$ preferences, $2$ traits,
and $4$ FoVs, e.g. $\{\mathrm{red},\mathrm{green},\mathrm{blue},\mathrm{yellow}\}$$\times\{\mathrm{random},\mathrm{hypo}\}\times\{3\times3,5\times5,7\times7,9\times9\}$.

\subsection{Actions and Intention Predictions}

\begin{figure}
\begin{centering}
\includegraphics[width=1\columnwidth]{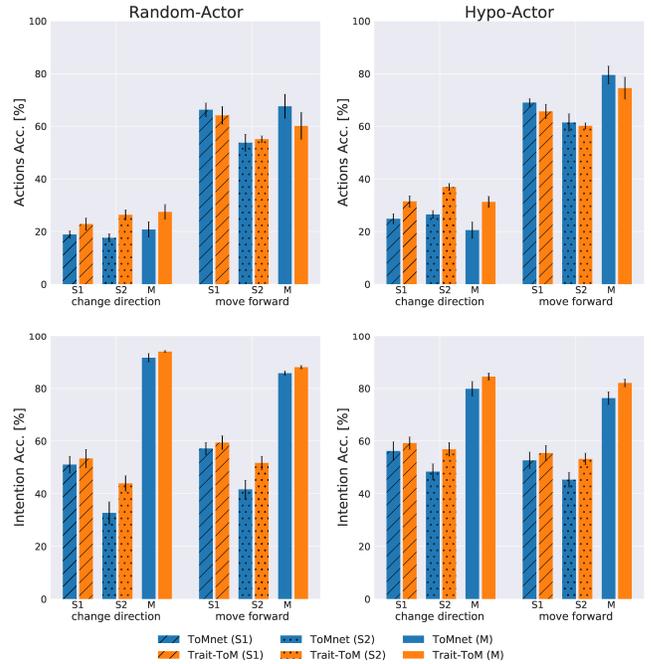}
\par\end{centering}
\caption{\label{fig_sequence_actors_move_actions}The y-axis shows the action
prediction accuracy (top row) and intention prediction accuracy (bottom
row) of ToMnet and Trait-ToM on the random-actor population (left
column) and the hypo-actor population (right column), conditioned
on two groups of move actions: (1) change direction ($\texttt{turn-left}$
and $\texttt{turn-right})$ and (2) move forward. Our Trait-ToM predicts
better when the actor change its direction.}
\end{figure}

\begin{table*}
\begin{centering}
\begin{tabular}{cccccccc}
\toprule 
\multirow{3}{*}{Stream} & \multirow{3}{*}{Model} & \multicolumn{2}{c}{Random-Actor} & \multicolumn{2}{c}{Hypo-Actor} & \multicolumn{2}{c}{Hypo-Actor}\tabularnewline
 &  & \multicolumn{2}{c}{(Full Observation)} & \multicolumn{2}{c}{(Full Observation)} & \multicolumn{2}{c}{(Partial Observation)}\tabularnewline
\cmidrule{3-8} \cmidrule{4-8} \cmidrule{5-8} \cmidrule{6-8} \cmidrule{7-8} \cmidrule{8-8} 
 &  & Action & Intention & Action & Intention & Action & Intention\tabularnewline
\midrule
\midrule 
\multirow{4}{*}{S1} & \multirow{2}{*}{ToMnet} & 37.80 & 53.21 & 54.26 & 53.43 & 49.65 & 51.94\tabularnewline
 &  & (0.64) & (2.83) & (0.94) & (3.18) & (0.96 ) & (2.71)\tabularnewline
\cmidrule{2-8} \cmidrule{3-8} \cmidrule{4-8} \cmidrule{5-8} \cmidrule{6-8} \cmidrule{7-8} \cmidrule{8-8} 
 & \multirow{2}{*}{Trait-ToM} & 39.46 & 55.64 & 54.37 & 56.45 & 50.93 & 55.21\tabularnewline
 &  & (0.47) & (3.02) & (1.09) & (2.48) & (1.10) & (2.43)\tabularnewline
\midrule 
\multirow{4}{*}{S2} & \multirow{2}{*}{ToMnet} & 32.48 & 36.60 & 50.29 & 46.59 & 44.30 & 43.96\tabularnewline
 &  & (1.71) & (3.84) & (2.32) & (2.67) & (2.45) & (3.09)\tabularnewline
\cmidrule{2-8} \cmidrule{3-8} \cmidrule{4-8} \cmidrule{5-8} \cmidrule{6-8} \cmidrule{7-8} \cmidrule{8-8} 
 & \multirow{2}{*}{Trait-ToM} & 38.22 & 47.26 & 52.85 & 54.60 & 48.80 & 53.32\tabularnewline
 &  & (1.25) & (2.77) & (0.81) & (2.35) & (1.20) & (2.30)\tabularnewline
\midrule 
\multirow{4}{*}{M} & \multirow{2}{*}{ToMnet} & 39.30 & 88.14 & 59.55 & 75.92 & 54.13 & 74.71\tabularnewline
 &  & (0.34) & (1.05) & (1.46) & (2.35) & (1.23) & (1.83)\tabularnewline
\cmidrule{2-8} \cmidrule{3-8} \cmidrule{4-8} \cmidrule{5-8} \cmidrule{6-8} \cmidrule{7-8} \cmidrule{8-8} 
 & \multirow{2}{*}{Trait-ToM} & \textbf{40.84} & \textbf{90.95} & \textbf{60.53} & \textbf{82.10} & \textbf{55.52} & \textbf{79.85}\tabularnewline
 &  & \textbf{(0.42)} & \textbf{(0.57)} & \textbf{(2.15)} & \textbf{(1.52)} & \textbf{(1.71)} & \textbf{(1.11)}\tabularnewline
\bottomrule
\end{tabular}
\par\end{centering}
\caption{\label{tab_action_intention_prediction} The accuracy of ToMnet and
Trait-ToM on predicting action and intention of (the $1^{st}$ and
$2^{nd}$ column) random-actors, hypo-actors in fully observable environments,
(the $3^{rd}$ column) hypo-actors in environments that have unobserved
obstacles to the observer (partial observation). Each cell contains
the mean (std.) over predictions of 6 runs.}
\vspace{-3mm}
\end{table*}
In the first experiment, we analyse the behaviour of ToMnet \citet{rabinowitz2018machine}
and our proposed Trait-ToM in predicting action and intention while
learning from the mixed and sequential population. In the mixed setting
(M), actors in one batch are i.i.d sampled from 32 actors. In sequential
settings (S), every $T_{stream}$, the observer sees a new type of
actors and never meets the previous actors again. The actors come
sequentially in increasing field of view order of \emph{hypo}- then
\emph{random}- actors (S1) or vice versa (S2). As a result, there
are totally three experimental settings in this task. Amongst all,
S1 and S2 are more realistic scenarios, e.g. the observer can only
see one type of actor at a time. Here, the observer can only learn
to predict behaviours of each type of actor for $T_{stream}=30,000$
iterations. The observer can see a batch of $B=16$ actors with the
same FoV and trait at each iteration. We assume that during the training
process, the actor will explicitly provide its intention ($\texttt{find}$,
$\texttt{goto}$, or $\texttt{pickup}$) to the observer as training
signals for the observer to predict its preferences, actions and intentions.
The actor reveals its individual characteristics such as trait and
FoV via its past behaviours ($N_{past}=3$).
\begin{figure}
\begin{centering}
\includegraphics[width=1\columnwidth]{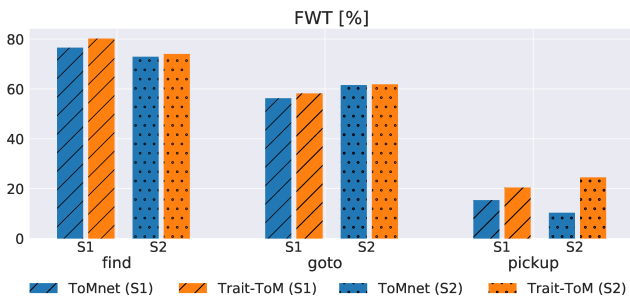}
\par\end{centering}
\caption{\label{fig_intent_fwt} Knowledge transfer in ToMnet and Trait-ToM
during the process of learning from different sequences of actors,
assessed by the positive forward transfer ability ($FWT$) on predicting
intention. Our Trait-ToM can learn patterns that are useful to predict
intention of different types of actors from the beginning of each
sequence (higher $FWT$).}

\centering{}\vspace{-3mm}
\end{figure}

\paragraph*{Results}

Fig.~\ref{fig_sequence_actors_prediction} shows the performance
of ToMnet and Trait-ToM on described tasks. To make a fair comparison
between the two methods, we designed two networks with roughly a similar
numbers of parameters (Trait-ToM has slightly less parameters than
ToMnet and they are different in the prediction networks). For each
architecture, we trained and reported the results over predictions
of 6 runs. In general, our trait-based ToM are able to predict the
behaviour of the actors better than ToMnet in all settings (M, S1
and S2). Both can predict precisely the preference of actors since
it is shown clearly in the past trajectories. Training observers on
the mixed population helps the models predict actor's intention $\texttt{find()}$
much better.

Fig.~\ref{fig_sequence_actors_move_actions} shows the action prediction
and intention prediction of ToMnet and Trait-ToM conditioned on move
actions. There are two groups of move actions: (1) change direction,
which includes $\texttt{turn-left}$ and $\texttt{turn-right}$; (2)
move forward. While both models can predict equally well when the
actors move forward, our Trait-ToM can predict better when the actors
will change their direction. Note that the moving forward action can
be easily predicted by looking at the current trajectory, however,
the change direction action depends on the individual characteristics
such as trait and field of view of the actors. This illustrates that
ToMnet heavily relies on the information of the current trajectory,
whereas our model uses the information revealed during past trajectories.

Table~\ref{tab_action_intention_prediction} summarises the action
and intention prediction accuracy of ToMnet and Trait-ToM. We also
evaluate both observers in predicting the behaviour of hypo-actors,
when both of them can only partially observe the environment. These
environments contain obstacles that are only observed by the actors.
As shown in Table~\ref{tab_action_intention_prediction}, the trait-based
observer outperforms the ToMnet in predicting the intention of actors.
Especially, our Trait-ToM, with the fast weight mechanism, outperforms
ToMnet by a larger margin when the actor streams are more realistic
(S1 and S2). 

\begin{figure*}
\begin{centering}
\includegraphics[width=0.7\textwidth]{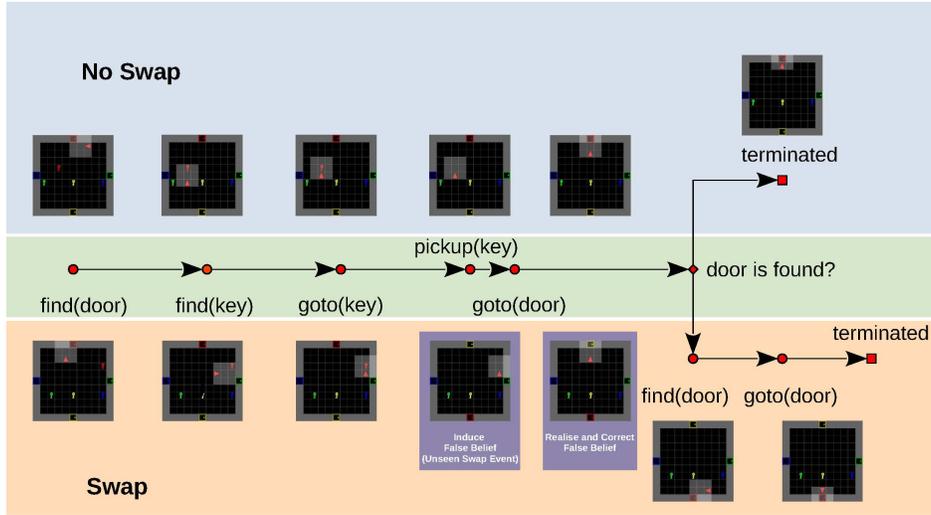}
\par\end{centering}
\caption{\label{fig_illu_key_door_env_hypo_agent}Dynamic intentions of hypo-actor
in \emph{key-door environment} with swap event (bottom) and no swap
event (top). The actor has a $3\text{x}3$ field of view and prefers
\textcolor{red}{red} colour. After seeing the \textcolor{red}{red}
door, the actor finds and collects the \textcolor{red}{red} key. The
actor then comes back to \textcolor{red}{red} door position that she
last saw and believes the \textcolor{red}{red} door is still there.
When there is no swap event, the actor successful reaches the door
(top). When there is a swap event, the actor realises that the \textcolor{red}{red}
door is not at the previous place, but a \textcolor{yellow}{yellow}
door instead. She changes her belief then tries to find and reach
the \textcolor{red}{red} door (bottom).}
\end{figure*}

To further understand the effect of the fast weight mechanism on the
learning process, we measure the knowledge transfer during this learning
process from two sequential settings (S1 and S2). Each stream will
provide for the training process a sequence $(\mathcal{A}_{i})_{i=1\dots M}$
of $M$ goal-directed actors which are different in their fields of
view and trait. Let $acc_{i,j}$ the observer's accuracy in predicting
the behaviour of actors $\mathcal{A}_{j}$ after being trained on
$(\mathcal{A}_{1},\dots,\mathcal{A}_{i})$. We then evaluate the forward
transfer ability defined as:
\[
FWT=\frac{\sum_{j=2}^{M}\sum_{i=1}^{j-1}acc_{i,j}}{\frac{1}{2}M(M-1)}.
\]
Intuitively, $FWT$ \citep{lesort2020continual} indicates how learning
to predict the behaviour of new actors affects the performance on
future unseen actors. Fig.~\ref{fig_intent_fwt} shows that our Trait-ToM
can learn useful knowledge to transfer and predict the intention of
actors sooner than ToMnet, as shown by a higher $FWT$.

\subsection{Direct Assessment of False-Belief Understanding }

The false belief in human can be assessed by various methods~\citep{beaudoin2020systematic}.
A common method to evaluate the computational theory of mind \citep{rabinowitz2018machine,nguyen2021theory}
is by constructing experiments of Sally-Anne Test - a classic false-belief
task \citep{wimmer1983beliefs,baron1985does}. In this task, the subject
observes a scene in which there are two dolls named by Sally and Anne.
Sally first puts her toy into the basket then goes out. While Sally
is outside, Anne takes the toy from Sally's basket and put in Anne's
box. The observer will be asked where will Sally finds the toy when
she comes back. Here we set up the key-door scenario with a swap event.
Fig.~\ref{fig_illu_key_door_env_hypo_agent} illustrates the trajectories
of an actor with a $3\times3$ field of view who prefers red colour
in the key-door environment with swap or no swap event. First, the
actor looks for the red door. After seeing it, the actor finds and
collects the red key. The actor then comes back to the red door position
and supposes it is still there. When there is no swap event, the actor
successfully reaches the red door. However, when there is a swap event,
the actor sees a yellow door instead and realises that the red door
has been swapped. She changes her belief then goes to find the red
door.

In this experiment, both ToMnet and Trait-ToM are trained on the mixed
population of hypo-actors and random-actors. The environment in which
actors operate may contain a swap event. In addition to the preference,
intention, and action prediction, the models are queried successor
representations to make a prediction about the long-term behaviours
of the actors. The ability to predict the difference in long-term
behaviours between the swap and non-swap events indicates the understanding
of the false belief.

\begin{figure}
\centering{}\includegraphics[width=1\columnwidth]{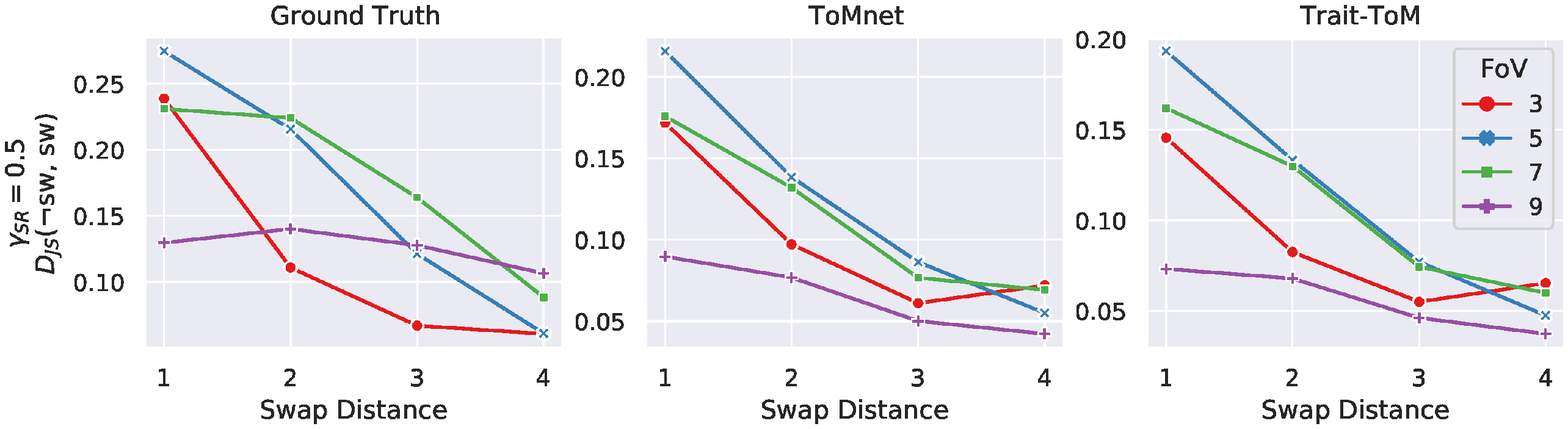}\caption{\label{fig_direct_false_belief} The Jensen--Shannon divergence between
successor representations when the actor picks up the key in swap
vs. no swap situation (left most) of the hypo-actor, (middle) predicted
by ToMnet, and (right most) predicted by Trait-ToM. The x-axis is
the distance from the preferred key to the target door. Statistical
analysis is shown in Table~\ref{tab_stats_direct_false_belief}.}
\end{figure}

\begin{table}
\begin{centering}
\begin{tabular}{|c|c|c|c|}
\hline 
 & $r_{\text{ToMnet}}$ & $r_{\text{Trait-ToM}}$ & $t$\tabularnewline
\hline 
\multirow{2}{*}{$3\times3$} & $0.929044$ & $0.935927$ & $2.12$\tabularnewline
 & $\pm0.00$ & $\pm0.00$ & $\pm3.42e\text{-}2$\tabularnewline
\hline 
\multirow{2}{*}{$5\times5$} & $0.911540$ & $0.931527$ & \multicolumn{1}{c|}{$6.22$}\tabularnewline
 & $\pm7.47e\text{-}298$ & $\pm0.00$ & $\pm8.34e\text{-}10$\tabularnewline
\hline 
\multirow{2}{*}{$7\times7$} & $0.893478$ & $0.910079$ & $5.39$\tabularnewline
 & $\pm1.62e\text{-}268$ & $\pm2.96e\text{-}295$ & $\pm9.16e\text{-}08$\tabularnewline
\hline 
\multirow{2}{*}{$9\times9$} & $0.849786$ & $0.864296$ & \multicolumn{1}{c|}{$3.14$}\tabularnewline
 & $\pm3.29e\text{-}215$ & $\pm8.18e\text{-}231$ & $\pm1.73e\text{-}03$\tabularnewline
\hline 
Conclusion & \multicolumn{2}{c|}{$>r_{0.05}(700)=0.074004$} & $>t_{0.05}(700)=1.64$\tabularnewline
\hline 
\end{tabular}
\par\end{centering}
\caption{\label{tab_stats_direct_false_belief}Statistical analysis of ToMnet
and Trait-ToM in predicting the false belief of actors with different
FoVs (row). The $r_{\text{ToMnet}}$ and $r_{\text{Trait-ToM}}$ are
Pearson correlation coefficients of ToMnet and Trait-ToM respectively.
The last column shows the $t\text{-}$test values computed by Steiger
method to compare two models based on the Pearson correlation coefficients.
Both models can predict $D_{JS}(\neg\text{sw},\text{sw})$. However,
Trait-ToM predicts closer to the ground truth than ToMnet.}
\end{table}

\begin{figure*}
\begin{centering}
\includegraphics[width=0.65\textwidth]{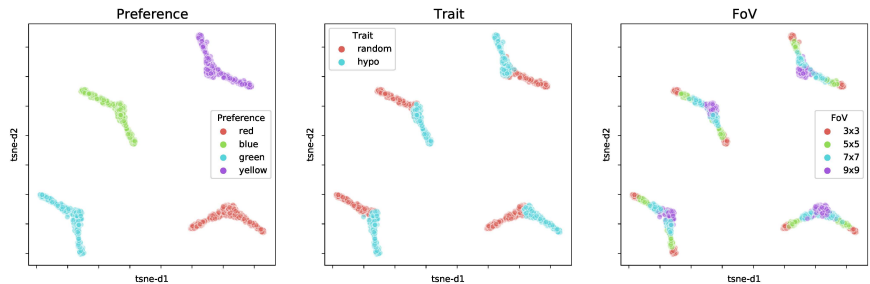}
\par\end{centering}
\caption{\label{fig_vis_tsne} Visualisation of weights of the prediction model
which is generated by the hypernetworks, projected into 2D using t-SNE.
The colours indicate the preferences (left), the traits (middle),
and the fields of view (right) of the actors. While the preferences
learning is supervised, the trait and field of view are learnt in
unsupervised manner. With the $9\times9$ field of view, both \emph{hypo-}
or \emph{random-} actors have similar behaviours, therefore, these
clusters are close (purple clusters at the right figure).}
\end{figure*}

Fig.~\ref{fig_direct_false_belief} shows the Jensen--Shannon divergence
between successor representations of the hypo-actor in swap versus
no swap environments of theory of mind models and the ground truth,
denoted as $D_{JS}(\neg\text{sw},\text{sw})$. In cases when the actor
can see the swap event, $D_{JS}(\neg\text{sw},\text{sw})$ is high
because the actor behaves differently from when there is no swap event.
As a result, when the swap distance increases, the behaviour of actors
with $9\times9$ fields of view (the purple line graph in Fig.~\ref{fig_direct_false_belief}
(left)) does not change much compared to actors with other fields
of view.

We calculate the Pearson correlation coefficients between the ground-truth
and the prediction of two models, shown in Table~\ref{tab_stats_direct_false_belief}.
Both methods can predict that there are differences in successor representations
of the \emph{hypo-}actor between the swap and no swap environments,
$r_{\text{ToMnet}}=0.929044\pm0.00>r_{0.05}(700)=0.074004$ and $r_{\text{Trait-ToM}}=0.935927\pm0.00>r_{0.05}(700)=0.074004$
for predicting actors with $3\times3$ field of view. To test whether
the predictions of Trait-ToM are significantly better than those of
ToMnet, we use Steiger method. This yields $t=2.12\pm3.42e\text{-}2>t_{0.05}(700)=1.64$,
hence the null hypothesis is rejected. We obtained similar conclusions
about the prediction of Trait-ToM and ToMnet on actors with other
FoVs (Table~\ref{tab_stats_direct_false_belief}).

\paragraph*{Visualisation}

We project the weight space of Trait-ToM into 2D using t-SNE to have
a better visual understanding our networks, c.f. Fig.~\ref{fig_vis_tsne}.
There are clear clusters of the preferences (left) confirming that
this information is explicitly coded in the training data. The Trait-ToM
produces different clusters given different traits (middle) and FoVs
of the actors (right) despite the fact that we do not train Trait-ToM
using this information. Especially, there is a smooth transition between
clusters of actors with different FoVs. We notice that random-actors
and hypo-actors with the furthest field of view ($9\times9$) in
this setting behave similarly; thus, the generated weights of our
hypernetworks for these types of actors form nearby clusters.

\subsection{Indirect Assessment of False-Belief Understanding }

Developmental psychologists indirectly assess the ability to understand
false belief by three experimental settings: (1) Violation of expectation
(VoE) \citep{onishi200515}; (2) Anticipatory looking \citep{clements1994implicit};
and (3) Active helping \citep{knudsen201218,buttelmann2009eighteen}.
These settings have inspired research in AI to construct benchmarks
to test the ToM models, e.g \citep{shu2021agent,gandhi2021baby} uses
VoE. We choose to indirectly assess our observer by the ability to
help other agents, i.e., the the active helping setting which the
belief attribution will appear within the context of intention attribution.
To pass this test, e.g. deliver proper helping behaviour, children
need to know that (1) others have goals, and at the same time (2)
others have false beliefs and thus may fail to achieve goals. Inspired
by this experiment, we implement a scenario in which an actor has
a false belief and an assistant needs to help with her (hidden) goal.
The closest setting to ours is \citep{puig2020watch}. However, there
the helper in that work only watches demonstrations to infer goals,
and their theory of mind model does not make any online prediction
about the behaviour of others. In our scenarios, we are able to investigate
action-based usages of theory of mind skills.

In this scenario, there are two agents: (1) the assistant (with or
without theory of mind) tries to assist, and (2) the actor who pursuits
its own goal. The assistant received full observations in the factored
representation from each environment and (optionally) predictions
from the theory of mind model. It can choose to give assistance directly
in action forms (the same as the actions which the actor can take
in the environment) or not to give any assistance and let the actor
act on its own. Note that in training RL agents, we assume any assistance
is costly, which means the agent needs to learn to take efficient
actions. The intention prediction is important in determining whether
to assist or not. A more realistic scenario in which there are more
than one agent acting on the same environment and the helper needs
to select who to help is left for future work. We especially highlight
here the scenario in which the actor can hold the false belief about
the position of the door. We first let the actor know the initial
position of the preferred door then create a swap event right after
the actor collects the key. The assistant needs to understand the
perspective of the actor to provide assistance. In addition, we consider
an \emph{obedient }actor which would follow any assistance if given
and act on its own strategy otherwise to achieve its goal.

\begin{algorithm} 
	\caption{The Procedure Help Policy } 
	\label{algo:ProcHelpPolicy}
	\SetKwInOut{Input}{Input}   
	\SetKwInOut{Output}{Output}
	\Input{ The theory of mind model ToM$(\cdot)$ \newline $\pi^*(a_t|s_t,g)$   the optimal policy}
	\Output{ The assistance given to the actor }  
	
	Predict the action of the actor $\tilde{a}_t, \tilde{g}\leftarrow ToM(s_t)$\;
	Compute the optimal action $\hat{a}_t\leftarrow \pi^*(a_t|s_t,\tilde{g})$\;
	\uIf{$\hat{a}_t = \tilde{a}_t$}{\textbf{return} no assistance\;
	}
	\Else{\textbf{return} $\tilde{a}_t$\;} 

\end{algorithm}

\begin{figure}
\begin{centering}
\includegraphics[width=1\columnwidth]{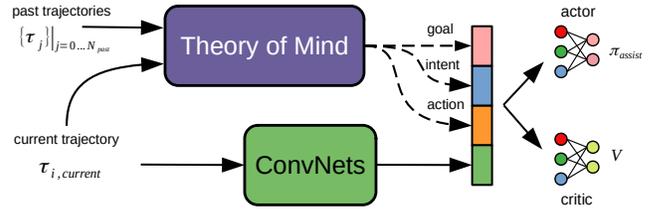} 
\par\end{centering}
\caption{\label{fig_rl_help_tom_policy}The architecture of the theory of mind
augmented reinforcement learner. The dash lines indicate there is
no gradient flow during training the assistance policy. The ToM model
is used as a forward model to generate goal, intention and action
predictions.}
\end{figure}

We construct two types of ToM-augmented help policy: (1) Procedure
Policy; and (2) Reinforcement Learner. The algorithm of the procedure
policy is shown in Algorithm~\ref{algo:ProcHelpPolicy}. To implement
the procedure policy, we need to pre-train the goal-conditioned policies
that can make near-optimal decisions based on full observations of
the environment $\pi^{*}(a_{t}|s_{t},g)$. The procedure policy will
give the near-optimal actions which is computed by $a_{t}^{*}=\pi^{*}(a_{t}|s_{t},g)$
as an assistance to the actor if the optimal action is different from
the predicted action of the ToM model. Our intuition is that if the
action of the actor is optimal, then it does not need to be assisted.
The constructions of reinforcement learner is shown in Fig.~\ref{fig_rl_help_tom_policy}.
Since the assistant can have access to full observations of the scene,
we do not need to use a memory mechanism such as one in LSTM or even
external memory to implement the policy. The reinforcement learning
(RL) agent assists the actor based on goal, action, and intention
predictions of the theory of mind model. We simply concatenate these
outputs of the theory of mind models to create the feature vector
of states observed from the environment. All information is given
to two heads, the actor and the critic, in order to predict the action
and the expected value, respectively. We trained the RL agents with
this actor-critic structure by Proximal Policy Optimisation (PPO)
algorithm \citep{schulman2017proximal}. 

\begin{figure}
\begin{centering}
\includegraphics[width=1\columnwidth]{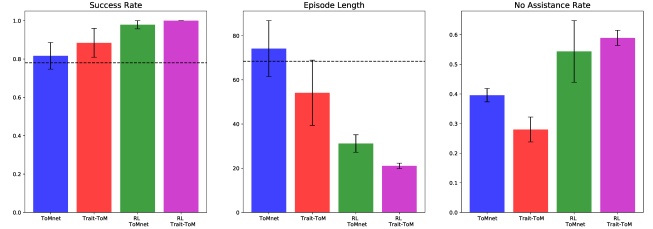} 
\par\end{centering}
\caption{\label{fig_indirect_false_belief_procedure_tom}Performance of ToM--augmented
assistants in helping \emph{hypo-}actor with the $3\text{x}3$ field
of view. The black dot line shows the performance of actors without
help. RL assistants assist better than procedure policy. Trait-ToM
RL assistants assists more efficiently than ToMnet RL assistants.}
\end{figure}

Fig.~\ref{fig_indirect_false_belief_procedure_tom} compares the
performance of ToM--augmented policies in the helping task. For each
ToM--augmented agent, we report the mean and standard deviation measures
over 6 runs. Since Trait-ToM can give better prediction for both type
of policies, it can help agents to achieve higher success rates than
ToMnet helped agents. Although both ToMnet and Trait-ToM can assist
actors achieve its task more frequent than acting on their own (all
bars higher than the horizontal black dot line in the most left figure),
Trait-ToM can help actors to complete tasks faster, i.e. the average
episode length is smaller (the middle figure). Comparing two groups
of assistants, the reinforcement learners perform better than the
procedure policies on this task, both in term of success rates and
times to completion. Interestingly, although the reinforcement learning
agent with Trait-ToM does not need to give as much assistance as ToMnet,
it still maintains higher success rates (the far left figure) and
lower time to achieve the task (the far right figure). This highlights
the importance of understanding false belief in helping other agents
where accurate false-belief understanding assists better and is more
efficient.

\section{Conclusions}

We have proposed a new Trait-based Theory of Mind (Trait-ToM) model
to equip social observers with the ability to infer the mental states
and goals of other actors through observing their past and current
behaviours. Central to our model is the idea that stable character
traits hold the key prior information that influences the transient
mental states. We hypothesise that such an influence has a multiplicative
nature -- traits may modulate the prediction path from the mental
states to the future behaviours. We realised these ideas through the
concept of `fast weights', in that the weights of the prediction
network are determined by the latent traits, which are functions of
the past behaviours, forming a \emph{hypernet} architecture for Trait-ToM.
We designed and conducted a suite of experiments over \emph{a key-door
environment}, in which actors have varied preference and intention
traits. The results showed that the multiplicative interactions between
the past and the present help make more accurate future predictions,
especially when trained in varying settings such as a mixed or sequential
population. Trait-ToM based assistant can also achieve a better performance
in providing helping\emph{} behaviours, known as indirect assessment
of \emph{false-belief} understanding.

 \balance

\bibliographystyle{ACM-Reference-Format}
\bibliography{main}

\end{document}